\documentclass[
 11pt]{article}
\usepackage[abbrvbib,preprint]{jmlr2e}
\usepackage[utf8]{inputenc}
\usepackage[T1]{fontenc}
\usepackage{algorithmic}
\usepackage{algorithm}
\usepackage{amsmath}
\usepackage{amsfonts}
\usepackage{hyperref}

\usepackage{tabularx}

  \makeatletter
  \@mparswitchfalse
  \makeatother
  \normalmarginpar 

    \usepackage[normalem]{ulem}

\newcommand{\bookbox}[1]{\small
\par\medskip\noindent
\framebox[\columnwidth]{
\begin{minipage}{0.8\minipagewidth} {#1} \end{minipage} } \par\medskip }

\newlength{\minipagewidth}
\newcommand{\be}{\begin{equation}}
\newcommand{\ee}{\end{equation}}

\newcommand{\Prob}[1]{\mathbb{P}\left(#1\right)}
\newcommand{\hmu}{\widehat{\mu}}
\newcommand{\De}{\Delta}
\newcommand{\hDe}{\widehat{\Delta}}
\newcommand{\hk}{\widehat{k}}
\newcommand{\E}{\mathcal{E}}

\newcommand{\eqdef}{\stackrel{\rm def}{=}}
\newcommand{\CR}[1]{R\left(#1\right)}
	\usepackage{lastpage}

\ShortHeadings{Semi-overlapping MAB best arm identification}{Antos et al}
\firstpageno{1}

\begin{document}

\title{Semi-overlapping multi-bandit best arm identification\\
 for sequential support network learning}

\author{\name András Antos \email antos@mit.bme.hu \\
       \addr Department of Artificial Intelligence and Systems Engineering\\
       Budapest University of Technology and Economics\\
       Budapest, Hungary
       \AND
       \name András Millinghoffer \email milli@mit.bme.hu \\
       \addr Department of Artificial Intelligence and Systems Engineering\\
       Budapest University of Technology and Economics\\
       Budapest, Hungary\\
       \addr E-Group ICT Software Zrt.\\
       Budapest, Hungary       
       \AND
       \name Péter Antal \email antal@mit.bme.hu \\
       \addr Department of Artificial Intelligence and Systems Engineering\\
       Budapest University of Technology and Economics\\
       Budapest, Hungary\\
       \addr E-Group ICT Software Zrt.\\
       Budapest, Hungary       
       }

\editor{}

\maketitle

\begin{abstract}
Many modern AI and ML problems require evaluating partners' contributions through shared yet asymmetric, computationally intensive processes and the simultaneous selection of the most beneficial candidates.
Sequential approaches to these problems can be unified under a new framework, \emph{Sequential Support Network Learning} (SSNL), in which the goal is to select the most beneficial candidate set of partners for all participants using trials;
 that is, to learn a directed graph that represents the highest-performing contributions.
 We demonstrate that a new pure-exploration model, the \emph{semi-overlapping multi-(multi-armed) bandit} (SOMMAB), in which a single evaluation provides distinct feedback to multiple bandits due to structural overlap among their arms, can be used to learn a support network from sparse candidate lists efficiently.

We develop a generalized GapE algorithm for SOMMABs and derive new exponential error bounds that improve the best known constant in the exponent for multi-bandit best-arm identification.
The bounds scale linearly with the degree of overlap, revealing significant sample-complexity gains arising from shared evaluations. 

From an application point of view, this work provides a theoretical foundation and improved performance guarantees for sequential learning tools for identifying support networks from sparse candidates in multiple learning problems, such as in multi-task learning (MTL), auxiliary task learning (ATL), federated learning (FL), and in multi-agent systems (MAS).
\end{abstract}


\section{Introduction}\label{sec:intro}

\emph{Multi-Armed Bandits} (MAB) is a widely used model today.
Its \emph{best arm identification} (BAI) setting \citep{audibert2010best,bubeck2011pure} is a \emph{pure exploration} problem, distinct from the original MAB problem to maximize the cumulative reward \citep[see e.g.,][]{Rob52,AuCeFi02finite}.

The core objective of BAI is to formulate a strategy that recommends the most beneficial one from $K$ possible options. 
The well-established MAB/BAI model has been motivated by the fact
 that the efficient allocation of trial resources is a critical aspect,
 given the significant computational, statistical, ethical, financial, and security/privacy-related budgetary limitations of trials in numerous domains.
The uniform exploration of options could result in the inefficient utilization of trial resources,
 potentially leading to suboptimal option selection.
Therefore, it is imperative to employ effective strategies for the distribution of trials across options,
 thereby ensuring the efficacy of the developed strategy.
To evaluate a MAB strategy, usually either the reward of the recommended arm or the probability of error (i.e., not selecting the best arm) is used.

To address the best arm identification problem, two algorithms were given by \citet{audibert2010best}:
 1) the \emph{Upper Confidence Bounds} (UCB-E) method with a parameter,
 whose optimal value depends on the complexity of the problem, and
 2) the parameter-free \emph{Successive Rejects} (SR) method.
Both these algorithms are shown to be nearly optimal,
 that is, their error probability decreases exponentially in the number of pulls.

BAI has been utilized, for example, in the context of function learning \citep{maron1993hoeffding,madani2004active,krueger2015fast,mohr2021lccv}; in the feature subset selection problem~\cite{gaudel2010feature,chaudhry2018feature}, and it is intensively used in hyper-parameter learning based on the results from random(ized) trials corresponding to train-test-validation splits~\cite{li2018hyperband}.
New class of applications of BAI in artificial intelligence are the selection of the best set of (1) jointly learnable tasks in multi-task learning (MTL) \citep{caruana1997multitask,standley2020tasks,fifty2021efficiently}, (2) beneficial tasks for a target task in auxiliary task learning (ATL) \citep{guo2019autosem,millinghoffer2024boosting}, (3) contributing clients in federated learning (FL) \citep{zhu2024client}, and (4) agents forming the most capable auxiliary coalition for an agent to assist in complex problem solving \citep{larsson2021automated,zhang2024coalitional,cohen2024online,cohen2025decentralized} (for a summary of application domains, see Table~\ref{tab:bai_applications}).
In these problems, the optimal option does not necessarily comprise all other entities (tasks, clients, or agents, respectively),
 because some of the entities may exert a deleterious effect on another, for instance, due to disparities in the sample distributions available for the entities.

\paragraph{Duality and the support network learning problem}
In these applications, another important property is the \emph{donor-recipient duality},
 in which entities serving as recipients (targets) may also act as donors (contributors) for others. Indeed, in many real-world problems, we may observe \emph{(complete) entity duality} denoting a setting in which all entities in the problem simultaneously play the role of both a donor and a recipient.
A further practically important property arises from the mandatory joint evaluation of participating entities, which implies computational and even implementation coupling.
The condition \emph{relational duality} formalizes the complete form of this property, which requires that if a set $S$ is considered as a candidate donor set for an entity $a$, then for any $b\in S$ the corresponding variant set $S\setminus \{b\} \cup \{a\}$ is available as a candidate donor set for \(b\),
 ensuring structural coherence of candidate relations across entities. By \emph{strong duality}, we mean the combination of entity and relational duality.
 This donor-recipient duality naturally leads to a joint selection problem, where we must determine the most beneficial set of donors for each (recipient) entity.
We refer to this problem class as \emph{sequential support network learning} (SSNL),
 where each entity aims to identify the most beneficial set of contributors using optionally computationally coupled trials in a sequential learning framework.
The solution for a SSNL problem can be represented by a directed graph with edges pointing to recipients from donors in the optimal support sets.

\paragraph{Towards multi-bandits} The support network learning problem can be viewed as a weakly coupled joint selection problem, which suggests using multi-bandits, a generalization of MABs, as follows.
We assume the existence of $M$ entities,
 each characterized by distinct joint learning properties
 and for entity $m$, $K_m$ candidates (options, arms) of beneficial entity subsets.
The objective is to identify the optimal option for each entity.
Certainly, if the number of options were allowed to be exponential in $M$ the problem easily becomes intractable;
 thus, in practice, that must be assumed to be limited.
We refer to this assumption as the candidate lists are \emph{sparse}.
Subsequently, it is necessary to solve $M$ MAB problems in parallel,
 while constrained by the total available resources.
It is possible that a greater quantity of resources will be required to identify the best option for one node than for another.
Therefore, employing a uniform or other arbitrary strategy generally will not result in optimal performance.
This problem structure is formulated by \citet{gabillon:hal-00632523,gabillon2011multi}
 as the \emph{multi-bandit/multi-MAB} (MMAB) best arm identification / pure exploration over $M$ MABs.

There are multiple ways to evaluate a MMAB strategy.
Three of them are the ones based on
\begin{enumerate}\setlength{\itemsep}{0pt plus 1pt minus 1pt} 
 \item the average of the rewards of the recommended arms over the bandits,
 \item the average error probability over the bandits,
 \item the maximum error probability over the bandits.
\end{enumerate}

The UCB-E and Successive Rejects algorithms above, being designed for a single MAB,
 are not obvious to extend to MMAB problems.
By \citet{gabillon:hal-00632523}, studying the MMAB problem in the fixed budget setting,
 the \emph{Gap-based Exploration} (GapE) algorithm has been proposed,
 which focuses on the \emph{gap} of the arm,
 that is, the difference between the mean value of the arm and that of the best arm (in the same bandit).
They prove an upper-bound on the error probability for GapE,
 which decreases exponentially with the budget (see Proposition~\ref{prop:GaGhLaBu11GapE_bound}),
 and they also report numerical simulations.

\begin{figure}[H]
    \centering
    \includegraphics[width=0.95\linewidth]{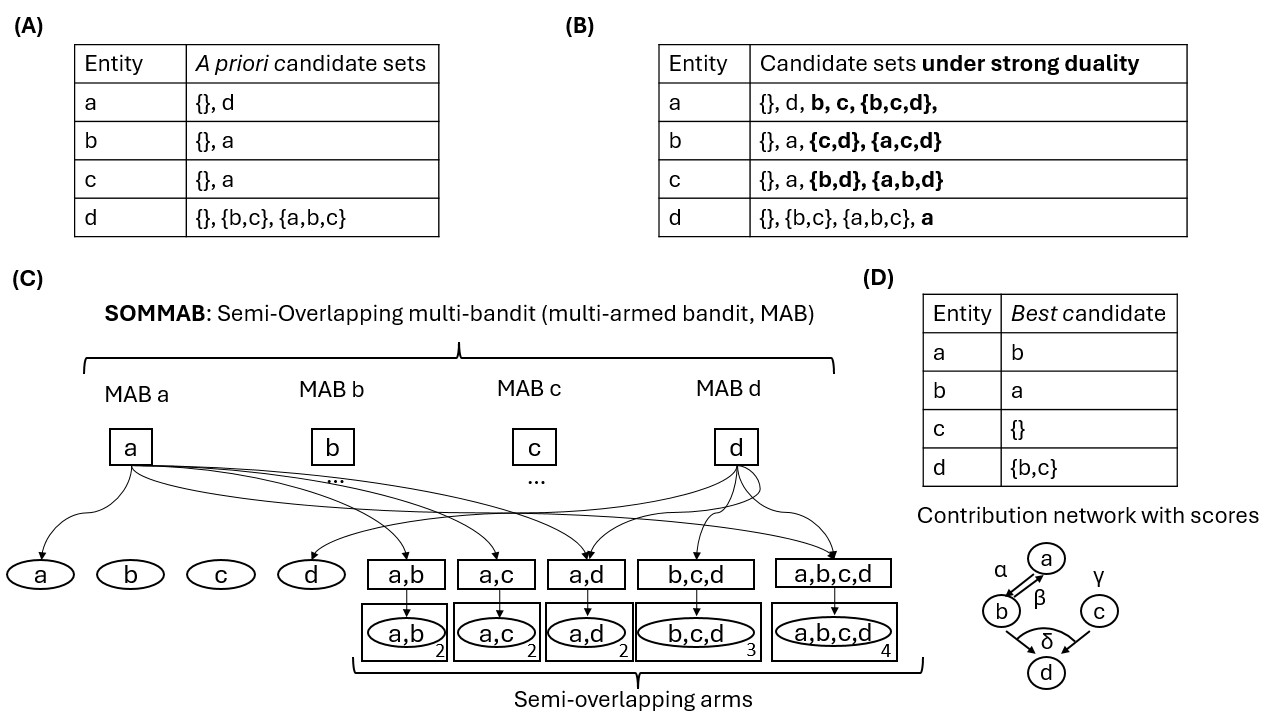}
    \caption{Overview of the Semi-Overlapping Multi-bandit (SOMMAB) formulation under donor--recipient duality.
\textbf{(A)} \emph{A priori candidate sets}:
 for each entity \(a,b,c,d\), we specify the initially admissible donor sets, which do not need to satisfy strong duality. 
In this example, the empty set is always included and the complete set is also a candidate for an entity.
\textbf{(B)} \emph{Candidate sets under strong duality}:
 enforcing complete entity duality and relational duality expands the candidate sets.
In particular, if a set is admissible for one entity, then for each of its members, the corresponding role-swapped variant is also admissible, yielding a structurally consistent family of candidate sets across all entities.
\textbf{(C)} \emph{SOMMAB}:
 each entity induces a local MAB problem, where arms correspond to its admissible candidate donor sets.
In a SOMMAB, arms may be \emph{semi-overlapping (SO)} across entities, as different MABs may mutually share arms corresponding to role-swapped variants of candidate sets.
The dual nature of SO arms is expressed using the plate notation: the cost of pulling a SO arm is a single cost for all corresponding MAB,
 whereas the random rewards are MAB-specific.
Synchronized sequential learning proceeds in parallel across MABs for the entities, with trials corresponding to selecting and evaluating candidate donor sets.
\textbf{(D)} \emph{Solution representation}:
 once learning converges, each entity selects its best-performing donor set, resulting in a directed support network.
Multi-node donor sets induce multiple incoming edges to a recipient,
 and sets of directed edges to a recipient represent learned contribution relationships.
This network compactly summarizes the outcome of the SOMMAB learning process.
Note that the support network may contain cycles, contribution scores can be asymmetric ($\alpha\neq \beta$),
 and contribution scores correspond to donor sets, as in the case of $\delta$ for the $\{b\rightarrow d, c\rightarrow d\}$ arc-connected edge set (or, with hypergraph representation, $\{b,c\}\rightarrow d$ hyperedge).}
    \label{fig:Overview4SOMMAB}
\end{figure}

\paragraph{From duality to the semi-overlapping property} The duality properties in support network learning formalized the underlying reality of joint computational evaluations in the learning process. Note that this does not mean hard constraints for the optimal support network. In multi-bandits the shared computations manifest as  \emph{computationally coupled arms}:
 evaluating the usefulness of auxiliary entities for a given recipient using a randomized trial, i.e., pulling an arm in one bandit,
 overlaps computationally with evaluating the usefulness of this recipient for the entities referred to as auxiliary, i.e., pulling an arm in another bandit.
On the other hand, the resulting contribution scores themselves (i.e., the output of these arms in distinct bandits) typically differ. Later we will formalize  these properties together  as the \emph{semi-overlapping} property of arms. In the pairwise case, it means that evaluating the usefulness of node $a$ for node $b$ shares computation with the evaluation of the usefulness of $b$ for $a$,
 even though these contribution scores can be asymmetric  (see Fig.~\ref{fig:Overview4SOMMAB}).

A key question in these diverse learning problems characterized by SSNL with strong duality is the following:\\
\emph{In order to identify the most beneficial candidate sets for all entities,
 how can we efficiently use sequential randomized computationally shared trials that provide per-entity feedback of the joint supports of the co-evaluated participants in the trial?}

In this paper,
 firstly we extend the MMAB model introducing the notion of \emph{semi-over\-lap\-ping arms set} motivated above,
 which consists of arms belonging to different bandits
 and overlapping in the sense that they are ``pulled together'',
 however, the reward distributions and the actual reward responses are not necessarily the same for these arms.
Then, we generalize the GapE algorithm and the corresponding upper-bound by \citet{gabillon:hal-00632523} mentioned above
 for this semi-overlapping setting,
 also improving the constant in the exponent by more than a factor of $3.5$,
 and by another factor of $r$ for $r$-order semi-overlapping multi-bandits (SOMMAB).

Organization of the paper:
 Section \ref{sec:related} summarizes the related works,
 Section~\ref{sec:MMABsetup} introduces formally the MMAB problem,
 Section~\ref{sec:gap_algo} describe the (generalized) GapE algorithm,
 Section~\ref{sec:error_bounds} gives the error probability bounds with a proof,
 which is also fixed, clarified and somewhat streamlined in comparison with the proof by \citet{gabillon:hal-00632523},
 Section~\ref{sec:ACFLappl} presents applications of SSNL,
 Section~\ref{sec:discussion} contains the discussion and conclusion, 
 Section~\ref{sec:future} summarizes possible future work,
 finally, some details of the proof are given in the Appendix.

\section{Related work}\label{sec:related} 

\paragraph{Multi-armed bandits and best-arm identification}
The classical MAB problem was introduced in the context of sequential experimental design by \citet{Rob52},
 with later developments focusing on regret minimization and upper confidence bound (UCB) strategies \citep{AuCeFi02finite}.
In contrast, BAI considers \emph{pure exploration},
 where the goal is to identify the optimal arm rather than maximize cumulative reward.
Fixed-budget and fixed-confidence formulations have been extensively studied,
 with algorithms such as UCB-E and SR shown to be at least nearly optimal (up to logarithmic factors in $K$) \citep{audibert2010best,bubeck2011pure}.
Furthermore, \citet{gabillon:hal-00747005} presented the UGapEb($0,1,n,a$) algorithm with also a parameter, whose optimal value depends on the complexity and with a bound that exhibits an exponent that is superior to that of the UCB-E algorithm.
Our work follows this pure-exploration line for MMABs.

\paragraph{Multi-bandits, overlapping and semi-overlapping (structured) MMABs}
Multi-bandit (or multi-task bandit) problems consider several bandits that must be solved in parallel under a shared sampling budget.
\citet{gabillon:hal-00632523,gabillon2011multi} introduced a multi-bandit BAI framework and the GapE algorithm, and established exponential error bounds governed by a global complexity measure aggregating per-bandit gaps.

A version of the MMAB setting, the (fully) \emph{overlapping multi-bandit} together with applications, was introduced by \citet{scarlett2019overlapping},
 where the same arm may belong to multiple bandits, thereby coupling the bandit problems.
When such an arm is pulled,
 it returns one reward response from its single reward distribution relevant in each of these bandits.
Here, we have to find the best arm in each bandit when arm sets may overlap.

Our SOMMAB model is closest in spirit to these works,
 but also allows for \emph{asymmetric} overlap:
 structurally linked arms may have distinct reward distributions, and a single evaluation provides different information to each bandit.
We show that this more general structure still admits GapE-type algorithms,
 and prove improved error exponents 
 exploiting structural overlap across bandits and analyzing how such structure affects the achievable error exponents that scale with the degree of overlap.

\paragraph{Combinatorial bandits and Monte Carlo Tree Search}
Combinatorial bandits provide a principled framework for subset-level decision making under limited evaluation budgets \citep{cesa2009combinatorial,chen2014combinatorial,rejwan2020top}.
In this setting, each action corresponds to selecting a subset (a “super-arm”) from a ground set, and the learner receives linear/additive feedback on the chosen subset — enabling gradual discovery of high-reward candidates via structured exploration and exploitation.

By contrast, Monte Carlo Tree Search (MCTS) offers a general-purpose heuristic for exploring exponentially large decision trees through repeated random simulations, combined with upper-confidence–based selection strategies (e.g., the UCT rule) to balance exploration and exploitation~\citep{KoSze:ECML06,browne2012survey,swiechowski2023monte}.
See, for example, applications of MCTS by~\citet{gaudel2010feature,chaudhry2018feature} in a closely related problem of feature subset selection (FSS).
Similar to MABs in FSS \citep{liu2021multi},
 our approach focuses on selection from an arbitrary, well-defined, sparse candidate list of contributor sets,
 allowing non-additive, interaction-driven reward structures.

\paragraph{Structure learning and sparse candidate methods}
In probabilistic graphical model learning, structure search over directed acyclic graphs is another instance of a high-dimensional, combinatorial optimization problem, although contrary to influence or causal diagrams, there is no 'hard' acyclicity constraint in the support network, only shared computational evaluations of candidates presents a 'soft' cost for the sequential learning process itself.
However, our approach shares the two-stage perspective of The Sparse Candidate method \citep{Friedman1999SparseCandidate}:
 it restricts the set of potential donors for each variable and then searches over this reduced space.

\paragraph{Computational efficiency of joint evaluations}
Computational aspects of joint evaluations have also motivated substantial work in multitask learning, such as the Task Affinity Grouping method ~\citep{fifty2021efficiently,millinghoffer2024boosting}.

\paragraph{Task selection in multi-task and auxiliary task learning}
Multi-task learning frequently grapples with negative transfer effects, that is, the intricate, complex pattern of beneficial-detrimental effects of certain tasks on others \citep{caruana1997multitask,xu2017demystifying,zhang2022survey}. It is by no surprise as the nature of transfer learning is multi-factorial and transfer effects can be attributed to (1) data representativity, (2) misaligned latent representations, (3) restrictive model learning capacity and (4) optimization bias;thus transfer effects are contextual and depend on the sample sizes, task similarities, sufficiency of hidden representations and model, and stages of the optimization \citep{ben2010theory,pentina2017multi,xin2022current}.
The auxiliary task learning approach suggests selecting the beneficial auxiliary task subsets for each target task \citep{guo2019autosem,shi2010auxiliary,standley2020tasks,fifty2021efficiently,song2022efficient,dery2022aang,jiang2024forkmerge,li2024barlow}.
However, evaluation of candidate auxiliary task subsets can be computationally demanding and leads to loss of statistical power,
 which could be mitigated by the application of MABs using task representations and carefully constructed candidate task subsets \citep{du2023multi,mukherjee2024multi,sessa2024multitask,millinghoffer2024boosting}. 

\paragraph{Client selection in federated learning}
Federated learning emerged from a distributed, yet centrally aggregated approach for learning from horizontally- and/or vertically-partitioned, but i.i.d.\ datasets \citep{wright2004privacy,ma2006privacy,yang2006privacy,samet2009privacy}.
Heterogeneity of partners in the mainstream formalization of federated learning became pivotal, and a large body of work has examined client selection and scheduling strategies to accelerate convergence, reduce communication, and improve fairness under heterogeneous data and system conditions \citep{liu2020client,lin2025communication,gupta2025federated,zhu2024client,ami2025client,trindade2024client,nagalapatti2021game,qu2022context}.
Recent hierarchical/personalized FL frameworks further emphasize structured interactions among clients~ \citep{huang2021compositional,tarzanagh2022federated,zhang2023fedaum,luo2023pgfed,shen2025multiobjectiveFL,shen2025surveyMOFL,zhang2025fedpref,wu2022pfedsv}.
Our work provides a complementary, pure-exploration viewpoint, in which client selection is reinterpreted as sequential identification of beneficial candidates under a shared evaluation budget.
Contrary to hierarchical FL, which aims to find, through a partitioning of partners defined by similar distributions, 'undirected', homogeneously beneficial coalitions for each coalition member, the SSNL formalization aims to find the most beneficial partner set for each participant.

\paragraph{Partner selection in multi-agent systems and multi-agent RL}
In multi-agent systems (MAS), coalition formation and partner selection have been extensively studied, often using cooperative game theory or learning-based mechanisms.
Recent work on cooperative MAS investigates how agents form, adapt, and evaluate coalitions under communication, incentive, and strategic constraints \citep{ota2022coalitional,jiang2022coalitionmass,chen2024freerider,chen2025dualgfl,chaudhury2022corefed,donahue2021optimality}.

These approaches typically assume symmetric or mutually agreed coalitions and often aim at stable or optimal coalition structures.
By contrast, SSNL explicitly models asymmetric donor–recipient relationships between partners, and SOMMAB fully represents and levarages the semi-overlapping evaluation pattern.
This shifts the focus from equilibrium coalition states to the \emph{sequential} identification of beneficial asymmetric coalitions under limited samples and computation, linking coalition formation in MAS with best-arm identification and structured pure exploration.

Finally, credit assignment learning in multi-agent reinforcement learning (MARL) provides a generalized, quantitative perspective to multi-bandit formulations by studying how multiple learning agents coordinate, compete, and transfer influence through structured interactions and shared environments \citep{hernandez2021survey}.
Recent MARL works on cooperative games, value decomposition, and population-based training naturally instantiate asymmetric contribution flows between agents, closely aligning with the support networks studied in this paper \citep{yang2020beyond,zhou2020learning,gorsane2022towards,bettini2024benchmarl}.

\paragraph{Dataset selection}

If, in federated learning (FL), clients each possess multiple local datasets, the partner selection problem becomes even more complex:
 rather than selecting a single client as a whole, the learning system must evaluate and choose among potentially many dataset contributions per client.
In such scenarios, each dataset offered by a client may vary in relevance, size, and distributional alignment with the global task, and the usefulness of a client is no longer a single scalar but a composition of contributions from its constituent datasets.
This exacerbates the selection challenge because a naive choice of clients may inadvertently favor those with many weak datasets over those with fewer but highly informative ones.
Formal dataset selection approaches from the broader machine learning literature,
 which frame dataset choice as an optimization problem over expected utility or distributional matching, provide valuable tools for quantifying dataset usefulness and can be integrated with client selection strategies to better balance contribution quality and quantity \citep{huang2023learning,fan2025assaymatch,xie2023data}.

\section{Multi-Bandit Problem Setup}\label{sec:MMABsetup}
 
Now we define the MMAB BAI problem more formally and introduce the notation used in this paper.
Let the number of MABs be denoted by $M$ and, for simplicity, the number of arms for each MAB by $K$.
The reward distribution of arm $k$ of bandit $m$ is denoted by $\nu_{mk}$.
It is bounded in $[0,b]$ and has mean $\mu_{mk}$.
Usually, $m$, $p$, or $q$ will denote MAB indices,
 and $k$, $i$, or $j$ denote arm indices.
It is assumed, that each bandit $m$ has a unique best arm $k_m^* := \arg\max_{1\le k\le K} \mu_{mk}$,
 and $\mu_m^* := \mu_{mk_m^*} = \max_{1\le k\le K} \mu_{mk}$ denotes the mean reward of $k_m^*$.
In bandit $m$, the gap for arm $k$ is defined as $\De_{mk} = |\max_{j\ne k} \mu_{mj}-\mu_{mk}|$.
That is, $\De_{mk} = \mu_m^*-\mu_{mk}$ for suboptimal arms,
 while the gap of $k_m^*$, $\De_{mk_m^*} = \mu_m^* - \max_{j\ne k_m^*} \mu_{mj}$ is the same as that of a second best arm.

The distributions $\{\nu_{mk}\}$ are not known to a forecaster,
 who, at each round $t=1,\ldots,n$, pulls a bandit-arm pair $I(t)=(m,k)$
 and observes a sample drawn from the distribution $\nu_{I(t)}$ independently from the past (given $I(t)$).
The $s^{\rm th}$ sample observed from $\nu_{mk}$ will be denoted by $X_{mk}(s)$,
 and the number of times that the pair $(m,k)$ has been pulled by the end of round $t$ is denote by $T_{mk}(t)$.
The forecaster estimates each $\mu_{mk}$ by computing the average of the samples observed over time,
 that is, by
\be\label{eq:def_hmu}
 \hmu_{mk}(t) = \frac1{T_{mk}(t)} \sum_{s=1}^{T_{mk}(t)} X_{mk}(s).
\ee
At the end of final round $n$,
 the forecaster specifies an arm with the highest estimated mean for each bandit $m$,
 that is, outputs $J_m(n) \in \arg\max_k \hmu_{mk}(n)$.
The error probability and the simple regret of recommending $J_m(n)$ is
 $e_m(n):=\Prob{J_m(n)\ne k_m^*}$ and $r_m(n):=\mu_m^*-\mu_{m J_m(n)}$, respectively.
Recalling the three performance measures mentioned in Sec.~\ref{sec:intro}, which are to be minimized,
the first one corresponds to the expectation (w.r.t.\ the samples) of the average regret incurred by the forecaster
\[
r(n) := \mathbb{E}\left[ \frac1{M} \sum_{m=1}^M r_m(n) \right] = \frac1{M} \sum_{m=1}^M \left( \mu_m^*-\mathbb{E}\mu_{m J_m(n)} \right),
\]
 the second one is the average error probability
\[
 e(n) := \frac1{M} \sum_{m=1}^M e_m(n) = \frac1{M} \sum_{m=1}^M \Prob{J_m(n)\ne k_m^*},
\]
 while the third one corresponds to the maximum (worst case) error probability
\[
 \ell(n) := \max_m e_m(n) = \max_m \Prob{J_m(n)\ne k_m^*}.
\]
\citet{gabillon:hal-00632523} gives the rationale of these measures in applications.
It is easy to see that the three measures are related through
\[
 \frac{\min_{m,k\ne k_m^*} \De_{mk}}{M} \cdot \ell(n) \le \min_{m,k\ne k_m^*} \De_{mk} \cdot e(n)
 \le r(n) \le b e(n) \le b \ell(n),
\]
 thus bounding one of these measures also gives control for the other two.
Thus the bounds both by \citet{gabillon:hal-00632523} and in this paper apply to $\ell(n)$.

In this paper,
 we extend MMABs allowing the bandits to have \emph{semi-overlapping arms}:
\begin{definition}\label{def:semi-overlapping}
A set of \emph{semi-overlapping} arms (i.e, an \emph{evaluation group}) consists of multiple arms belonging to different bandits that are ``overlapping'' in the sense that they are ``pulled together'',
 that is, their joint trial consumes one pull from the budget,
 however, the reward distributions and the actual reward responses are not necessarily the same for these arms.
Here it is assumed for simplicity that such sets are pairwise disjoint.

An MMAB with semi-overlapping arms is a \emph{semi-overlapping multi-bandit} (SOMMAB).

For any $r\ge 2$ integer, an \emph{$r$-order SOMMAB} is one,
 where each arm belongs to a set of semi-overlapping arms and
 the size of each maximal set of semi-overlapping arms is at least $r$.
Moreover, we can call all MMAB a $1$-order semi-overlapping MMAB.
\end{definition}

Note that when a set of semi-overlapping arms is pulled
 then for all arms in the set, the corresponding $T_{mk}(t)$'s are increased simultaneously,
 that is, in case of semi-overlapping, the sum of $T_{mk}(t)$'s may exceed $t$.
Namely, for $r$-order semi-overlapping multi-bandit, the sum of $T_{mk}(t)$ is at least $rt$.

Also note that for SSNL,
 the different sets of semi-overlapping arms are disjoint and correspond to the sets of nodes trialed together.
Thus if always at least $r$ entities are trialed together,
 that is, each candidate contributor set has at least size $r-1$,
 then the SSNL problem corresponds to an $r$-order SOMMAB.
Therefore, we call it an \emph{$r$-order SSNL}. 

\section{The GapE Algorithm}\label{sec:gap_algo}

Using the definition of $\hmu_{mk}(t)$, the estimated gaps are defined in an analogous way as $\De_{mk}$:
\be\label{eq:def_hDelta}
 \hDe_{mk}(t) = |\max_{j\ne k} \hmu_{mj}(t)-\hmu_{mk}(t)|.
\ee
We have slightly generalized the Gap-based Exploration (GapE) algorithm by \citet{gabillon:hal-00632523}.
In particular, instead of one pull for each arm in the initialization phase,
 the algorithm takes a new parameter $l$,
 and it applies $l$ initializing pulls per arm.
Also, to make it aligned with SOMMABs,
 we detailed how GapE should handle semi-overlapping arms if any.
The clarified pseudo-code of this generalized GapE(l) algorithm is shown in Figure~\ref{alg:GapE-algo}.
\begin{figure}[htb]
\bookbox{
\begin{algorithmic}
\STATE \textbf{Parameters:} number of rounds $n$, exploration parameter $a$, maximum range $b$, uniform exploration length $l$
\STATE \textbf{Initialize:}
draw each bandit-arm pair $(m,k)$ * $l$-times\\
\qquad for $t=lMK$ set all $T_{mk}(t)=l$, calculate $\hmu_{mk}(t)$ as in \eqref{eq:def_hmu}
\STATE \qquad\qquad\qquad\qquad calculate all $\hDe_{mk}(t)$ as in \eqref{eq:def_hDelta}
\FOR{$t=1,2,\ldots,n$}
  \STATE Compute $B_{mk}(t) = -\hDe_{mk}(t-1) + b\sqrt{\frac{a}{T_{mk}(t-1)}}$ for all bandit-arm pairs $(m,k)$
  \STATE Draw $I(t)\in\arg\max_{mk} B_{mk}(t)$ *
  \STATE Observe $X_{I(t)} (T_{I(t)}(t-1)+1) \sim \nu_{I(t)}$
  \STATE Update $T_{I(t)}(t)=T_{I(t)}(t-1)+1$ and $\hmu_{I(t)}(t)$ as in \eqref{eq:def_hmu}
  \STATE Update all $\hDe_{mk}(t)$ as in \eqref{eq:def_hDelta} for all bandit involved by semi-overlapping
  \ENDFOR
\STATE Return: $J_m(n)\in\arg\max_{k\in\{1,\ldots,K\}} \hmu_{mk}(n)$, $\forall m\in\{1,\ldots,M\}$
\STATE
\STATE * Whenever the drawn arm is part of an evaluation group,
 $I(t)$ should represent the entire group, that is,
 all arms in it receive a reward and their corresponding $T_{mk}(t)$, $\hmu_{mk}(t)$, and all affected $\hDe_{mk}(t)$ are updated.
\end{algorithmic}
}
\caption{The pseudo-code of the generalized Gap-based Exploration (GapE(l)) algorithm.}
\label{alg:GapE-algo}
\end{figure}

Based on the estimated gaps instead of the means,
 GapE handles the $M$-bandit problem somewhat like a single-bandit problem with $MK$ arms is handled by UCB-E.
At time step $t$,
 an index $B_{mk}(t)$ is computed for each pair $(m,k)$ by GapE based on the samples up to time $t-1$.
Then the pair $(m,k)$ with the highest $B_{mk}(t)$ is selected as $I(t)$ (together with arms in the same semi-overlapping set as $I(t)$, if any).
Similarly to UCB methods of \citet{AuCeFi02finite}, $B_{mk}$ has two terms.
The first one is the negative of the estimated gap for the bandit-arm pair (since the gap is to be minimized),
 while the second one is a confidence radius based exploration term
 pushing GapE to pull arms that have been underexplored.
This way,
 GapE tends to balance between pulling arms with less estimated gap and smaller number of samples.
The exploration tendency of GapE can be tuned by the exploration parameter $a$.

When the time horizon $n$ is known,
 both Proposition~\ref{prop:GaGhLaBu11GapE_bound} and Theorem~\ref{th:GapE_bound} in Section~\ref{sec:error_bounds}
 suggest that $a$ should be set as $a = C \tfrac{n-MK}{H}$,
 where $C$ stands for appropriate constants specified in the results, respectively, and
\[
 H = \sum_{m,k} \frac{b^2}{\De_{mk}^2}
\]
is the complexity of the MMAB problem introduced by \citet{gabillon:hal-00632523}
 (see further discussion by \citealp[Section~3.1]{gabillon:hal-00632523} and in Section~\ref{sec:error_bounds} below).
As pointed out by \citet[Section~3]{gabillon:hal-00632523},
 this algorithm differs from most usual bandit strategies in that the index $B_{mk}$ for one arm depends explicitly on the samples from the other arms, as well.
Due to this, the analysis of GapE is much more complicated.
See also the thorough comparison of the GapE and UCB-E algorithms and
 for an explanation on why the gaps are crucial instead of means in the multi-bandit case by \citet[Section~3]{gabillon:hal-00632523}.
        
Note that
 the idea of GapE is extended by \citet[Section~3.2]{gabillon:hal-00632523} also into a more involved algorithm, \emph{GapE-variance} (GapE-V),
 which takes into account the variances of the arms, as well,
 and exponential upper-bound on GapE-V's error probability is also provided.
 
Since the good range of the exploration parameter $a$ in both GapE and GapE-V
 depends on the problem complexity $H$, which is usually unknown in advance,
 adaptive versions of GapE (somewhat similarly to Adaptive UCB-E of \citealp{audibert2010best}) and GapE-V are also given there.
Moreover,
 the performance of GapE, GapE-V and their adaptive versions are
 evaluated by \citet[Section~4, Appendix~A]{gabillon:hal-00632523} on several synthetic problems,
 which demonstrates that
 these algorihms outperform benchmark strategies (\emph{Uniform}, \emph{Uniform+UCB-E}).

\section{Error probability bounds}\label{sec:error_bounds}

First we present the upper-bound on the error probability $\ell(n)$ of GapE by \citet[Section~3.1]{gabillon:hal-00632523} for (non-overlapping) MMABs:
\begin{proposition}\label{prop:GaGhLaBu11GapE_bound}
If we run GapE with parameters $l=1$ and $0<a \le 4\frac{n-MK}{9H}$, then its probability of error satisfies
\[
 \ell(n)
 \le \Prob{\exists m: J_m(n) \ne k_m^*}
 \le 2MKn \exp\left(-\frac{a}{64}\right),
\]
 in particular for $a = 4\frac{n-MK}{9H}$, we have
 $\ell(n) \le 2MKn \exp\left(-\frac{n-MK}{144H}\right)$.
\end{proposition}

\citet{gabillon:hal-00632523} give also some remarks about the bound above,
 the relation of the complexity $H$ to the complexity $H_m$ of each individual bandit and to the complexity $H_{mk}$ of each individual arm in each bandit,
 the comparison of \emph{GapE} with the \emph{static} allocation strategy
 (achieving at most $MK\exp(-n/H)$ error probability),
 and with the \emph{Uniform} and combined \emph{Uniform+UCB-E} allocation strategies
 (achieving at most $MK\exp(-n/(MK\max_{m,k} H_{mk}))$ and $2MKn\exp(-(n-MK)/(18M\max_m H_m))$ error probability, respectively).

As stated by \citet{gabillon:hal-00632523}: ``the constants in the analysis are not accurate.''
In this paper,   
 we could improve the constant in the exponent above by more than a factor of $3.5$ (for $l=152$, and by more than a factor of $2.4$ for $l=1$)
 taking it much closer to the constant $1/18$ in the bound for the single-MAB algorithm UCB-E by \citet{audibert2010best},
 and by another factor of $r$ for $r$-order semi-overlapping multi-bandits:
\begin{theorem}\label{th:GapE_bound}
If we run GapE with parameters $1\le l\le 152$ and $0<a \le \frac{n-MK+1}{(1+2c)^2H-Q_c}$ for a SOMMAB,
where $Q_c\eqdef 3(1+5c)(1+c)/4$,
\[
 c \eqdef \frac1{2\sqrt{3\rho+\rho^2}+2\rho+1} \qquad\mbox{ and }\qquad
 \rho\eqdef \sqrt{\min\left(\frac{l}{l-1},2\right)},
\]
 then for $n\ge lMK$, its probability of error satisfies
\[
 \ell(n)
 \le \Prob{\exists m: J_m(n) \ne k_m^*}
 \le 2MKn e^{-2ac^2},
\]
 in particular for $a = \frac{n-MK+1}{(1+2c)^2H-Q_c}$, we have
\[
 \ell(n) \le 2MKn \exp\left(-\frac{2(n-MK+1)}{(1/c+2)^2H-Q_c/c^2}\right).
\]
Especially,
\[
 \ell(n) < 2MKn \exp\left(-\frac{n-MK+1}{41H-36}\right),\qquad\mbox{if $l=152$.}
\]

For $r$-order semi-overlapping MMABs,
 if we run GapE with $a = \frac{rn-MK+1}{(1+2c)^2H-Q_c}$ and $l=152$, we have
\[
 \ell(n)
 \le 2MKn \exp\left(-\frac{2(rn-MK+1)}{(1/c+2)^2H-Q_c/c^2}\right)
 < 2MKn \exp\left(-\frac{rn-MK+1}{41H-36}\right).
\]
\end{theorem}
\begin{remark}[Choice of $l$]
Despite the apparent complexity of the formulae defining $c$,
 it is actually a rather flat function of $l$ taking values in the interval $(1/9,1/7)$.
For the natural $l=1$ choice,
 one could use, for example, simply $c=6/53$ instead,
 when Theorem~\ref{th:GapE_bound} implies for $n\ge MK$,
\[
 \ell(n) < 2MKn \exp\left(-\frac{n-MK+1}{59H-50}\right).
\]
In this investigation, however, our objective was to identify the optimal exponent achievable with this proof technique by adjusting $l$.
Nonetheless, the optimal selection of $l=152$ may initially appear enigmatic.
This choice is presumably an artifact stemming from the involved technique.
On the one hand, as $l/(l-1)$, and thus $\rho$ approaches $1$,
 the exponent that could be achieved in the bound decreases.
On the other hand, when $l$ is excessively large, the base step of the induction in the proof is invalid.
It is evident that the assumption of an unlimited $l$ is not a reasonable one,
 as it would result in a uniform arm sampling instead of any bandit strategy.
The choice of $l$ that yields the true optimal minimax rate is not necessarily identical to the above choice.
\end{remark}

\begin{example}[Comparison]
It is our understanding that Theorem~\ref{th:GapE_bound} leads to the best available bound for the MMAB problem.%
\footnote{The UGapEb algorithm for single MABs is extended to MMABs by \citet[Appendix~B]{gabillon:hal-00747005},
 and it is mentioned that their bound could be stated similarly for MMABs.
However, upon communication with the authors, it was revealed that
 no formal proof exists;
 only intuition for possibly extending the single MAB proof for MMABs.}
To compare this bound to the ones by \citet{gabillon:hal-00632523},
 as an example, examine a realistic situation,
 where the maximum of the complexities is $5$ times their average,
 that is, $M\max_m H_m/H \approx 5$.
Then, for the exponents,\\
* the bound for Uniform+UCB-E yields $-(n-MK)/90H$,\\
* the GapE bound by \citet{gabillon:hal-00632523} yields the weeker $-(n-MK)/144H$, while\\
* our new GapE bound yields $-(n-MK)/59H$ ($l=1$),
 or even $-(n-MK)/41H$ (for $l=152$ and $n\ge 152MK$).\\
Also, in the setting, where $M=K=2$, $H=25$ and $n=20000$,
 the GapE bound by \citet{gabillon:hal-00632523} yields the useless $160000 \exp(-4999/900)=61923\%$,
 while our new GapE bound yields $160000 \exp(-19997/1425)=12.873\%$ ($l=1$),
 or even $160000 \exp(-19997/989)=0.02648\%$ ($l=152$).

Moreover, numerical simulations by \citet{gabillon:hal-00632523} suggest that
 the optimal value of $aH/n$ is typically in range $[2,8]$.
Observe, that for large $n$, Proposition~\ref{prop:GaGhLaBu11GapE_bound} allows $aH/n$ to be increased up to $0.44$,
 whereas Theorem~\ref{th:GapE_bound} allows it up to $0.66$ for $l=1$.
While this suggest an improving trend, also indicates room for further improvement.
\end{example}
\begin{remark}
For (massively overlapping) SOMMABs,
 the number $n$ of pulls
 in the exponent is somewhat weak, 
 and instead we could expect intuitively the total sample size $\sum_{m,k} T_{mk}(n)$
 there.
However,
 this sum is not known in advance, so we cannot set the parameter $a$ depending on it.
On the other hand,
 the best lower bound for this sum that is known in advance, is $rn$ for $r$-order SOMMABs,
 and that is exactly what we have at the end of Theorem~\ref{th:GapE_bound}.
\end{remark}
\begin{remark}[$M$-order semi-overlapping]\label{rem:M-orderSO} 

For the $M$-order semi-overlapping case,
 where the pull of each of the $K$ semi-overlapping arm sets (recall that these are disjoint),
 actually allows a trial in each bandit,
 the bound takes the form
 $\ell(n) < 2MKn \exp\left(-\frac{n-K}{41H/M}\right)$,
 with the complexity being the average complexity over the bandits.
This makes sense considering that this configuration is equivalent to a single MAB with $K$ arms that possess $M$-dimensional vector values, where we have to find the best values dimension-wise.
See Section~\ref{sec:ACFLappl} for the relevance of approximately $M$-order semi-overlapping for SSNL.
\end{remark}

We have also fixed some flaws, clarified numerous details and slightly streamlined presentation of the original proof by \citet{gabillon:hal-00632523}:

\begin{proof}[of Theorem~\ref{th:GapE_bound}.]
If $2ac^2\le \log(2MKn)$, the bound on the probability is trivial.
If the Theorem holds for $b=1$,
 then the algorithm fed by the normed inputs $X_{mk}(t)/b$ and run by $b=1$ would make the same draws and same returns,
 so the Theorem follows for any $b$.
Thus w.l.o.g.\ we assume $2ac^2>\log(2MKn)$ and $b=1$.

Define $\CR{t}\eqdef\sqrt{a/t}$ for $t>0$,
 and recall that
 \[
  I(t)\in\arg\max_{mk} \left( -\hDe_{mk}(t-1) + \CR{T_{mk}(t-1)} \right).
\]

\smallskip{\bf Step 1.}
Let us consider the following event:
\[
 \E =
 \left\{ \forall m\in\{1,\ldots,M\}, \forall k\in\{1,\ldots,K\}, \forall t\in\{1,\ldots,n\},
 \left|\hmu_{mk}(t)-\mu_{mk}\right|<c\CR{T_{mk}(t)} \right\}
\]
From Chernoff-Hoeffding's inequality and the union bound, we have
\[
 \Prob{\xi^C} \le 2MKn e^{-2ac^2}.
\]
Now we would like to prove that on $\E$,
 we find the best arm for all the bandits, i.e., $J_m(n) = k_m^*$, $\forall m\in\{1,\ldots,M\}$.
All statements below are meant on $\E$.
Since $J_m(n)$ is the empirical best arm of bandit $m$,
 we should prove that for any $k'\in\{1,\ldots,K\}$, $k'\ne k_m^*$ implies $\hmu_{mk'}(n) < \hmu_{m k_m^*}(n)$.
This follows if we prove that for any $(m,k)$, $c\CR{T_{mk}(n)} \le \De_{mk}/2$,
 since then upper- and lower-bounding the $\hmu$'s by definition of $\E$,
\begin{align*}
 \hmu_{m k_m^*}(n) - \hmu_{mk'}(n)
& > \mu_{m k_m^*} - c\CR{T_{m k_m^*}(n)} - \mu_{mk'} - c\CR{T_{mk'}(n)}\\
&\ge \De_{mk'} - \De_{m k_m^*}/2 - \De_{mk'}/2\\
&\ge \De_{mk'} - \De_{m k'}/2 - \De_{mk'}/2
 = 0.
\end{align*}
Moreover, the desired $c\CR{T_{mk}(n)} \le \De_{mk}/2$ is equivalent to $T_{mk}(n) \ge \frac{4ac^2}{\De_{mk}^2}$.

\smallskip{\bf Step 2.}
In this step, we show that in GapE, for any bandits $(m,q)$ and arms $(k,j)$, and for any $t\ge lMK$,
 the following relation between the number of pulls of the arms holds
\be\label{eq:Step2_induction_ineq}
 -\De_{mk} + (1+2c)\CR{\max(T_{mk}(t)-1,1)} \ge -\De_{qj} + \frac{1-c}{2}\CR{T_{qj}(t)}.
\ee
We prove this inequality by induction on $t$.

\smallskip\emph{Base step.}
Recall that $\rho=\sqrt{\min(l/(l-1),2)} \in (1,\sqrt{2}]$ for $l\ge 1$
 and $c = \tfrac1{2\sqrt{3\rho+\rho^2}+2\rho+1} \in (1/9,1/7)$,
 and that
\be\label{eq:a_lowerbound}
 2ac^2>\log(2MKn) \ge \log(2MKMKl) \ge \log(32l)
\ee
 for $n\ge lMK$, $M,K\ge 2$.
We know that after the first $lMK$ rounds of the GapE algorithm,
 all the arms have been pulled $l$-times, i.e., $T_{mk}(lMK)=l, \forall m,k$.
Thus, for $t=lMK$, inequality \eqref{eq:Step2_induction_ineq} is equivalent to
\be\label{eq:gen_basestep_ineq}
 \frac{(2+4c)\rho-1+c}{2}\sqrt{\frac{a}{l}} =
 \left(\frac{1+2c}{\sqrt{l-1}}-\frac{1-c}{2\sqrt{l}}\right)\sqrt{a}
 \ge \De_{mk}-\De_{qj}
 \qquad\mbox{for $l\ge 2$},
\ee
 and to
\[
 \frac{1+5c}2\sqrt{a}
 \ge \De_{mk}-\De_{qj}
 \qquad\mbox{for $l=1$}.
\]
The latter holds, since \eqref{eq:a_lowerbound} yields $\sqrt{a}c > \sqrt{\tfrac52\log 2}$ if $l=1$,
 and thus $\frac{1+5c}2\sqrt{a} > \tfrac52\sqrt{\tfrac52\log 2} > 1 \ge \De_{mk}-\De_{qj}$.

For the $l\ge 2$ case,
 we prove in Appendix~\ref{sec:app_proof_thm_GapE_bound} the following
\begin{lemma}\label{le:rho_ineq}
For $\rho>1$, $2\rho-1 > (11-4\rho)c$.
\end{lemma}

The inequality in this lemma can be rearranged as
 $(2+4c)\rho-1+c > 12c$,
 leading to
\[
 \frac{(2+4c)\rho-1+c}2 \sqrt{\frac{a}{l}} > 6c\sqrt{\frac{a}{l}}.
\]
Since, by \eqref{eq:a_lowerbound}, $36ac^2 > 18\log(32l) > l$ for $l\le 152$,
 we have $6c\sqrt{a/l}>1$,
 which, together with $\De_{mk}-\De_{qj}\le 1$, implies \eqref{eq:gen_basestep_ineq}.

\smallskip\emph{Inductive step.}
Let us assume that \eqref{eq:Step2_induction_ineq} holds at time $t-1$ and we pull arm $i$ of bandit $p$ at time $t$, i.e., $I(t)=(p,i)$ and $T_{pi}(t)\ge l+1$.
So at time $t$, the inequality \eqref{eq:Step2_induction_ineq} trivially holds for every choice of $m$, $q$, $k$, and $j$,
 except when $(m,k)=(p,i)\ne(q,j)$.
As a result, in the inductive step,
 we only need to prove that the following holds for any $q\in\{1,\ldots,M\}$ and $j\in\{1,\ldots,K\}$
\be\label{eq:induction_ineq_pi}
 -\De_{pi} + (1+2c)\CR{T_{pi}(t)-1}
 = -\De_{pi} + (1+2c)\CR{\max(T_{pi}(t)-1,1)}
 \ge -\De_{qj} + (1-c)\CR{T_{qj}(t)}/2.
\ee
 for $(q,j)\ne(p,i)$.
Since arm $i$ of bandit $p$ has been pulled at time $t$, we have that for any bandit-arm pair $(q,j)$
\be\label{eq:ineq4}
 -\hDe_{pi}(t-1)+\CR{T_{pi}(t-1)} \ge -\hDe_{qj}(t-1)+\CR{T_{qj}(t-1)}.
\ee
To prove \eqref{eq:induction_ineq_pi},
 we first prove an upper-bound for $-\hDe_{pi}(t-1)$ and a lower-bound for $-\hDe_{qj}(t-1)$:
\begin{lemma}\label{le:hDe_bounds_by_De}
For $c\le 1/3$,
\be\label{eq:ineq5}
 -\hDe_{pi}(t-1) \le -\De_{pi} + 2c \CR{T_{pi}(t)-1}
 \quad \text{ and } \quad
 -\hDe_{qj}(t-1) \ge -\De_{qj} - \left(2\frac{1+2c}{1-c}\rho+1\right)c\CR{T_{qj}(t)}.
\ee
\end{lemma}
We report the proofs of Lemma~\ref{le:hDe_bounds_by_De} in Appendix~\ref{sec:app_proof_thm_GapE_bound}.
The inequality \eqref{eq:induction_ineq_pi}, and as a result, the inductive step is proved
 by replacing $-\hDe_{pi}(t-1)$ and $-\hDe_{qj}(t-1)$ in \eqref{eq:ineq4} from \eqref{eq:ineq5}
 and under the condition that
\[
 \left(2\frac{1+2c}{1-c}\rho+1\right)c\le \frac{1+c}2,
 \qquad\mbox{ or equivivalenly }\qquad
 (8\rho-1)c^2+(4\rho+2)c-1
 \le 0.
\]
This condition is satisfied, since $c$ is just the higher root of the polynomial above.

\smallskip{\bf Step 3.}
In order to prove the condition of $T_{mk}(n)$ in Step~1,
 let us assume that arm $k$ of bandit $m$ has been pulled less than $\frac{a(1-c)^2}{4\De_{mk}^2}$ times at time $t=n$ (at the end),
 which is equvivalent to $-\De_{mk} + \frac{1-c}2 \CR{T_{mk}(n)} > 0$.
From this inequality and \eqref{eq:Step2_induction_ineq}, we have $-\De_{qj} + (1+2c) \CR{T_{qj}(n)-1} > 0$,
 or equivalently $T_{qj}(n) < \frac{a(1+2c)^2}{\De_{qj}^2}+1$ for any pair $(q,j)$.
We also know that
 $\sum_{q,j} T_{qj}(n) \ge n$ (with equality if there is no semi-overlapping).
From these, taking the assumption on $T_{mk}(n)$ and $\De_{mk}^2\le 1$ into account, we deduce that
\[
 n-(MK-1)
 < \sum_{q,j} \frac{a(1+2c)^2}{\De_{qj}^2} -\frac{a(1+2c)^2}{\De_{mk}^2} +\frac{a(1-c)^2}{4\De_{mk}^2}
 = a(1+2c)^2 H - \frac{aQ_c}{\De_{mk}^2}
 \le ((1+2c)^2 H - Q_c)a,
\]
where $Q_c=3(1+5c)(1+c)/4$.
So, if we select $a$ such that $n-MK+1 \ge ((1+2c)^2H-Q_c)a$,
 we have a contradiction,
refuting the assumption above that $T_{mk}(n) < \frac{a(1-c)^2}{4\De_{mk}^2}$,
 which means that $T_{mk}(n) \ge \frac{4ac^2}{\De_{mk}^2}$ for any pair $(m,k)$, when $c\le 1/5$.
The condition for $a$ in the theorem comes from our choice of $a$ above in this step.

For an $r$-order semi-overlapping multi-bandit, we also know that $\sum_{q,j} T_{qj}(n) \ge rn$,
 thus the inequality above becomes
\[
 rn-(MK-1) < ((1+2c)^2H-Q_c)a,
\]
 so if we choose $a = \frac{rn-MK+1}{(1+2c)^2H-Q_c}$,
 we must have the desired $T_{mk}(n) > \frac{4ac^2}{\De_{mk}^2}$ for any pair $(m,k)$,
 yielding
 \[
 \ell(n)
 \le 2MKn \exp\left(-\frac{2(rn-MK+1)}{(1/c+2)^2H-Q_c/c^2}\right)
 < 2MKn \exp\left(-\frac{rn-MK+1}{41H-36}\right)
\]
for $l=152$.
This concludes the proof.
\end{proof}

\section{Applications of sequential support network learning}\label{sec:ACFLappl} 

The SSNL appears in many recent applications, such as task selection in auxiliary task learning, client selection in federated learning, and agent selection in multi-agent systems (see Table~\ref{tab:bai_applications})..

\begin{table}[H]
\centering
\caption{Sequential support network learning in practice: entities, relationships, and trial mechanisms in various application domains.}
\label{tab:bai_applications}
\begin{tabularx}{\linewidth}{l l l X}
\textbf{Problem} & \textbf{Entity} & \textbf{Option} & \textbf{Trial} \\
Multi-Task Learning
& Task 
& Multi-output model
& Empirical estimate of composite error using random re-sampling\\

Auxiliary Task Learning
& Task 
& Auxiliary task set 
& Empirical estimate of target task error using random re-sampling \\

Federated Learning 
& Client 
& Collaborating clients
& Empirical estimate of target client error using random re-sampling \\

Multi-Agent Systems
& Agent
& Agent coalition
& Empirical estimate of target agent's score using randomized  problem solving \\
\end{tabularx}
\end{table}

In these applications, a usual heuristic for the construction of candidate sets is based on top-down approaches,
 that is, filtering entities with detrimental effects.
Consequently, the size of each candidate set, and thus the order of the SSNL, can be close to the number $M$ of entities.
For such scenarios, the following consequence of Theorem~\ref{th:GapE_bound}
is especially useful:
\begin{corollary}
If in $r$-order SSNL with $M$ entities,
 we run GapE with $a = \frac{rn-MK+1}{(1+2c)^2H-Q_c}$ and $l=152$, we have
\[
 \ell(n)
 < 2MKn \exp\left(-\frac{rn-MK+1}{41H-36}\right),
\]
 where $K$ is the (maximal) number of candidates for a node.
\end{corollary} 

Thus when $r$ is close to $M$, in line with Remark~\ref{rem:M-orderSO},
 the exponent of the bound can take the form
 $\frac{rn/M-K}{41H/M} \approx \frac{n-K}{41H/M}$
with the complexity being the average complexity over the entities.

For the specific applications above, the SSNL realization and MMAB model are detailed as follows.

\subsection{Multi-task learning and auxiliary task learning}

In the MTL/ATL approach, tasks correspond to bandits,
 arms are a priori defined candidate auxiliary task sets for each task,
 and pulls are evaluations of the performance of the joint model, including the target task and the auxiliary task set corresponding to the pulled arm, which evaluation may use a ``$k$-fold'' cross-validation.

\subsection{Federated learning}

In the FL framework, the multi-bandit model is applied as follows:
 separate bandits correspond to  clients or partners in FL,
 arms are a priori defined candidate partners sets for each partners,
 and pulls are standard FL evaluations using ``k-fold'' cross-validation and multiple hypothesis testing correction.
This application also utilizes the client-edge-cloud compute continuum \citep{liu2020client,lin2025communication,gupta2025federated}: the cloud server at a higher layer can run the MMAB algorithm to orchestrate the joint learning of auxiliary partner sets for each partner, 
 whereas edge nodes at lower layers can do the evaluations by running the standard elementary FL schemes for each partner.

\subsection{Coalition Learning in Multi-agent systems}

SSNL is also close to the learning of cooperations in \emph{multi-agent systems}, optimizing statistical, computational, and communication aspects \citep{ota2022coalitional,jiang2022coalitionmass,chen2024freerider,chen2025dualgfl,chaudhury2022corefed}.
However, in our approach, auxiliary partner selection is asymmetric, resembling the selection of an optimal tool set for an agent, and each partner can separately select its auxiliary partner set to improve its performance \citep{larsson2021automated,zhang2024coalitional,cohen2024online,cohen2025decentralized};
 see game-theoretic analysis of dependencies by \citet{donahue2021optimality}.

The multi-bandit model is applied as follows:
 agents correspond to bandits,
 arms are a priori defined candidate agent sets for each agent,
 and pulls are evaluations of the joint cooperation (problem solving) with stochastic elements.

\section{Discussion and Conclusion}\label{sec:discussion}

We demonstrated that a broad class of modern learning problems, where entities contribute to and benefit from others, and where contributions can be explored through weakly coupled shared sequential evaluations, can be cast as instances of a newly introduced framework, \emph{Sequential support network Learning} (SSNL).
Conceptually, SSNL concerns the identification and evaluation of beneficial auxiliary interactions among entities and can be viewed as learning a directed graph encoding the most supportive relations.

We showed that the SSNL problem can be solved efficiently by formulating it as a \emph{semi-overlapping multi-bandit} (SOMMAB) pure-exploration problem.
In this model, each candidate set corresponds to an arm, and structurally coupled evaluations induce semi-overlapping pulls in which a single evaluation provides distinct information to multiple bandits.
This abstraction captures the essential asymmetry and shared computational structure of SSNL while retaining the tractability of classical best-arm identification.

Building on this foundation, we established new exponential error bounds for SOMMABs under a generalized GapE algorithm.
The bounds significantly improve the constants known for non-overlapping multi-bandit BAI and scale linearly with the degree of overlap.
These results show that even partial structural coupling across bandits can meaningfully reduce sample complexity, providing theoretical justification for exploiting shared evaluations in distributed learning systems.

We further outlined realizations of the SOMMAB abstraction in several prominent learning paradigms---including multi-task learning (MTL), auxiliary task learning (ATL), federated learning (FL), and coalition formation in multi-agent systems (MAS).
These instantiations illustrate that semi-overlapping evaluation is a recurring structural feature across disparate machine learning applications.
In particular, the SSNL–SOMMAB formulation yields a new architectural lens for federated learning that aligns naturally with emerging cloud–edge infrastructures:
 the SOMMAB algorithms provide a principled mechanism for cloud-level orchestration of the exploration of auxiliary collaborations across clients with evaluations at the edge layers.

\section{Future Work} \label{sec:future}

Several directions emerge naturally from this work. Immediate theoretical challenges are as follows:

\begin{itemize}
   \item \textbf{GapE with variance} A variant of GapE, the GapE-V algorithm taking into account the variances of the arms
 is also proposed by \citet{gabillon:hal-00632523},
 along with a corresponding bound,
 which may be improved and extended to semi-overlapping case similarly as Proposition~\ref{prop:GaGhLaBu11GapE_bound}.
    \item \textbf{Adaptive algorithms} Adaptive versions of both GapE and GapE-V are also proposed by \citet{gabillon:hal-00632523} without proofs,
 where the complexity measures of the multi-bandit problem are estimated on the fly, so they do not need to be known in advance for tuning the exploration parameter $a$.
Giving bounds for the performance of these adaptive variants is a challenging problem.

    \item \textbf{$(\epsilon,m)$-best arm identification} Two common generalizations of the MAB/BAI problem are as follows:\\
- Rather than identifying the optimal arm, for given $\epsilon$,
 the objective is to identify an arm whose mean is closer than $\epsilon$ to that of the optimal one.\\
- For given $m$, the objective is to identify the best $m$ arm, rather than the single best.\\
The combination of these two generalizations gives rise to the $(\epsilon,m)$-best arm identification problem, as introduced by \citet{gabillon:hal-00632523},
 and for which GapE and Theorem~\ref{th:GapE_bound} may be extended.
\end{itemize}

Extending SOMMAB beyond fixed sparse candidate sets is another important avenue.
One natural extension is to remove the fixed-candidate constraint and move toward open-ended coalition optimization.
This requires navigating a combinatorial space of exponentially many coalitions.
We propose to incorporate \emph{Monte Carlo Tree Search (MCTS)}~\citep{KoSze:ECML06,gaudel2010feature,chaudhry2018feature,browne2012survey} as a meta-level selection mechanism over the support network.

Additional open problems include incorporating stochastic or adversarial overlap patterns, handling partial observability across bandits, and integrating privacy or communication constraints that arise in distributed or federated environments.
Each of these generalizations would further broaden the applicability of SSNL/SOMMABs and deepen the theoretical foundations of structured pure-exploration algorithms for cooperative agent learning.

\acks{
This research was supported by the National Research, Development, and Innovation Fund of Hungary under Grant OTKA-K139330, TKP2021-EGA-02, and the European Union project RRF-2.3.1-21-2022-00004 within the framework of the Artificial Intelligence National Laboratory.
}

\appendix
\section{Proofs of Lemmata in Theorem~\ref{th:GapE_bound}}\label{sec:app_proof_thm_GapE_bound}

In this appendix we prove the lemmata in the proof of Theorem~\ref{th:GapE_bound}:

\begin{proof}[of Lemma~\ref{le:rho_ineq}.]
Substituting $c>0$, it is to show that 
\[
 (2\rho-1)(2\sqrt{3\rho+\rho^2}+2\rho+1) = (2\rho-1)/c > 11-4\rho,
\]
 or equivivalently,
\[
 (2\rho-1)\sqrt{3\rho+\rho^2} > 6-2\rho-2\rho^2.
\]
This would follow from
\[
 (2\rho-1)\sqrt{3\rho+\rho^2} > |6-2\rho-2\rho^2|,
\]
 which, when squared, becomes
\[
 (2\rho-1)^2(3\rho+\rho^2) > (6-2\rho-2\rho^2)^2.
\]
After some algebra, this is equivalent to
$(\rho+4)(\rho-1)
 >0$,
 which clearly holds for $\rho>1$.
\end{proof}

\begin{proof}[of Lemma~\ref{le:hDe_bounds_by_De}.]
We denote by $\hmu_m^*(t)$ and $\hk_m^*(t)$ the estimated mean and the index of an (arbitrarily chosen) expirically best arms of bandit $m$ after round $t$
 (i.e., $\hmu_m^*(t) = \hmu_{m \hk_m^*(t)}(t) = \max_{1\le k\le K} \hmu_{mk}(t)$, $\hk_m^*(t) \in \arg\max_{1\le k\le K} \hmu_{mk}(t)$).
Also, by $\mu_m^+=\mu_{m k_m^+}$ and $k_m^+$ the mean and the index of an (arbitrarily chosen) second best arms of bandit $m$,
 and by $\hmu_m^+(t)=\hmu_{m \hk_m^+(t)}(t)$ and $\hk_m^+(t)$ the estimated mean and the index of an (arbitrarily chosen) empirically second best arms of bandit $m$ after round $t$.
(If the empirically best arm is not unique
 then the empirically best arms and the empirically second best arms constitute the same set,
 and $\hmu_{m \hk_m^+(t)}(t)=\hmu_{m \hk_m^*(t)}(t)$,
 however, $\hk_m^+(t)\ne \hk_m^*(t)$ is chosen.
Note that $\hmu_{m \hk_m^*(t)}(t)$, $\mu_{m k_m^+}$, and $\hmu_{m \hk_m^+(t)}(t)$ are unambiguous,
 but $\mu_{m \hk_m^*(t)}$, $\hmu_{m k_m^+}(t)$, and $\mu_{m \hk_m^+(t)}$ may be arbitrarily chosen from multiple values.)

\subsection*{Upper Bound in \eqref{eq:ineq5}}
Here we prove that
\[
 -\hDe_{pi}(t-1) \le -\De_{pi} + 2c\CR{T_{pi}(t)-1},
\]
 where arm $i$ of bandit $p$ is the arm pulled at time $t$.
This means that $T_{pi}(t-1) = T_{pi}(t)-1$.
Each
 $T$,
 $\mu$, $\hmu$,
 $\De$, $\hDe$, $k^*$, $\hk^*$, $k^+$ and $\hk^+$
refers for bandit $p$ below,
 so we supress the indices $p$ for these values here.
Also, each
 $T$, $\hmu$, $\hDe$, $\hk^*$ and $\hk^+$
refers to time $t-1$ below,
 so we omit the argument $t-1$ for these values here.
Thus, we prove $\hDe_i \ge \De_i - 2c\CR{T_i}$.
Observe that, since arm $i$ of bandit $p$ is pulled at time $t$, from \eqref{eq:ineq4} we have all of
\begin{align}
  -\hDe_i + \CR{T_i} &\ge -\hDe_{\hk^+} + \CR{T_{\hk^+}},\label{eq:algo_pulls_i_hk+}\\
 -\hDe_i + \CR{T_i} &\ge -\hDe_{\hk^*} + \CR{T_{\hk^*}}, \mbox{ and}\label{eq:algo_pulls_i_hk*}\\
 -\hDe_i + \CR{T_i} &\ge -\hDe_{k^*} + \CR{T_{k^*}}\label{eq:algo_pulls_i_k*}.
\end{align}
We consider the following four cases and use $0\le c\le 1$ throughout:

\medskip{\bf Case 1.}
$i=\hk^*$ and $i=k^*$:

Now we have
\begin{align*}
 \hDe_i
&= \hmu_i-\hmu_{\hk^+}\\
&\ge \mu_i-\mu_{\hk^+} -c\CR{T_{\hk^+}}-c\CR{T_i}\\
&\stackrel{(a)}{\ge} \mu_i-\mu_{\hk^+} -2c\CR{T_i}\\
&\ge \mu_i-\mu_{k^+} - 2c\CR{T_i}
 = \De_i - 2c\CR{T_i}.
\end{align*}

{\bf (a)}
By definition $\hDe_i = \hDe_{\hk^+}$,
 thus \eqref{eq:algo_pulls_i_hk+} gives $\CR{T_i} \ge \CR{T_{\hk^+}}$.

\medskip{\bf Case 2.}
$i=\hk^*$ and $i\ne k^*$:
Now we have $\hDe_i\ge 0$ obviously, while
\begin{align*}
 \De_i-2c\CR{T_i}
& = (1-c)(\De_i - c\CR{T_i}) + c(\De_i - (1+c)\CR{T_i})\\
&\stackrel{(b)}{\le} (1-c)(\De_i - c\CR{T_i}) - c(1-c)\CR{T_{k^*}}\\
&= (1-c)(\mu_{k^*} - \mu_i - c\CR{T_i} - c\CR{T_{k^*}})\\
&\le (1-c)(\hmu_{k^*} - \hmu_{\hk^*})\qquad\mbox{(on $\E$)}\\
&\le 0.\qquad\mbox{(by definition of $\hmu_{\hk^*}$)}.
\end{align*}
{\bf (b)} is proven using \eqref{eq:algo_pulls_i_k*} and $i=\hk^*\ne k^*$, implying
\[
 \hmu_i + \CR{T_i}
 \ge \hmu_{\hk^+} + \CR{T_i}
 \ge \hmu_{k^*} + \CR{T_{k^*}},
\]
which, using bounds on $\E$, further implies
\[
 \mu_i+(1+c)\CR{T_i} \ge \mu_{k^*}+(1-c)\CR{T_{k^*}}.
\]
Rearranging $\De_i - (1+c)\CR{T_i} \le -(1-c)\CR{T_{k^*}}$.

\medskip{\bf Case 3.}
$i\ne\hk^*$ and $i=k^*$:

Now we have
\begin{align*}
 \hDe_i
&= \hmu_{\hk^*}-\hmu_i\\
&\ge \hmu_{k^*}-\hmu_{\hk^*}\\
&\ge \mu_{k^*}-\mu_{\hk^*}-c\CR{T_{\hk^*}}-c\CR{T_i}\\
 &\stackrel{(c)}{\ge} \mu_{k^*}-\mu_{k^+}-c\CR{T_i}-c\CR{T_i}
 = \De_i-2c\CR{T_i}.
\end{align*}

{\bf (c)}
By definition $\hDe_i \ge \hDe_{\hk^*}$,
 thus \eqref{eq:algo_pulls_i_hk*} gives $\CR{T_i} \ge \CR{T_{\hk^*}}$.

\medskip{\bf Case 4.}
$i\ne\hk^*$ and $i\ne k^*$:
Now we have
\be\label{eq:ineq9}
 \hDe_i
 = \hmu_{\hk^*}-\hmu_i
 \ge \hmu_{k^*}-\mu_i - c\CR{T_i}
 \ge \mu_{k^*}-\mu_i - c\CR{T_i} - c\CR{T_{k^*}}.
\ee
Recall that by \eqref{eq:algo_pulls_i_k*}
\be\label{eq:ineq10}
 \hmu_i-\hmu_{\hk^*} + \CR{T_i} \ge -\hDe_{k^*} + \CR{T_{k^*}}.
\ee

If $k^*=\hk^*$, \eqref{eq:ineq10} can be written as
\[
 \hmu_i + \CR{T_i}
 \ge \hmu_{\hk^+} + \CR{T_{k^*}}.
\]
By $i\ne\hk^*$ and definition of $\hk^+$, $\hmu_{\hk^+} \ge \hmu_i$,
 which gives $\CR{T_i} \ge \CR{T_{k^*}}$ and thus $\hDe_i \ge \De_i - 2c\CR{T_i}$ from \eqref{eq:ineq9}.

If $k^*\ne\hk^*$, \eqref{eq:ineq10} can be written as
\[
 \hmu_i + \CR{T_i} \ge \hmu_{k^*} + \CR{T_{k^*}}
\]
which, using bounds on $\E$, implies
\[
 \mu_i + (1+c)\CR{T_i} \ge \mu_{k^*} + (1-c)\CR{T_{k^*}},
\]
and rearranging
 $-\CR{T_{k^*}} \ge \frac{\De_i}{1-c} - \frac{1+c}{1-c}\CR{T_i}$.
Plugging this into \eqref{eq:ineq9} we have
\[
 \hDe_i
 \ge \De_i - c\CR{T_i} + \frac{c\De_i}{1-c} - c\frac{1+c}{1-c}\CR{T_i}
 = \frac{\De_i - 2c\CR{T_i}}{1-c}.
\]
Now if $\De_i-2c\CR{T_i}\ge 0$ then the r.h.s.\ is lower bounded by $\De_i-2c\CR{T_i}$ as required.
If $\De_i-2c\CR{T_i}<0$, we are done due to $\hDe_i\ge 0$.

\subsection*{Lower Bound in \eqref{eq:ineq5}}

Here we prove that
\[
-\hDe_{qj}(t-1) \ge -\De_{qj} - (2\rho(1+2c)/(1-c)+1)c\CR{T_{qj}(t)}
\]
 for all bandits $q\in\{1,\ldots,M\}$ and all arms $j \in\{1,\ldots,K\}$,
 such that the arm $j$ of bandit $q$ is not the one pulled at time $t$, i.e., $(q,j)\ne (p,i)$.
This means that $T_{qj}(t-1) = T_{qj}(t)$.

In a manner akin to the proof for the upper-bound in Part~1, each
 $T$,
 $\mu$, $\hmu$,
 $\De$, $\hDe$, $k^*$, $\hk^*$, $k^+$ and $\hk^+$
refers for bandit $q$ below,
 so we supress the indices $q$ for these values here.
Also, each
 $T$, $\hmu$, $\hDe$, $\hk^*$ and $\hk^+$
refers to time $t-1$ below,
 so we omit the argument $t-1$ for these values here.
Thus, we prove
 $\hDe_j \le \De_j + (2\rho(1+2c)/(1-c)+1)c\CR{T_j}$.
Observe that, from the inductive assumption, we have all of
\begin{align}
 -\De_j + (1+2c)\CR{\max(T_j-1,1)} &\ge -\De_{k^+} + (1-c)\CR{T_{k^+}}/2,\label{eq:induction_i_k+}\\
 -\De_j + (1+2c)\CR{\max(T_j-1,1)} &\ge -\De_{k^*} + (1-c)\CR{T_{k^*}}/2,\mbox{ and}\label{eq:induction_i_k*}\\
 -\De_j + (1+2c)\CR{\max(T_j-1,1)} &\ge -\De_{\hk^*} + (1-c)\CR{T_{\hk^*}}/2.\label{eq:induction_i_hk*}
\end{align}
We consider the following four cases:

\medskip{\bf Case 1.}
$j=\hk^*$ and $j=k^*$:

Now we have
\begin{align*}
 \hDe_{j}
& = \hmu_j-\hmu_{\hk^+}\\
&\le \mu_j-\hmu_{k^+} + c\CR{T_j}\\
&\le \mu_j-\mu_{k^+} + c\CR{T_{k^+}} + c\CR{T_j}\\
&\stackrel{(e)}{\le} \De_j + 2c(1+2c)\rho\CR{T_j}/(1-c) + c\CR{T_j}
 = \De_j + (2\rho(1+2c)/(1-c)+1)c\CR{T_j}.
\end{align*}

{\bf (e)}
By definition $\De_{k^+} = \De_j$,
 thus \eqref{eq:induction_i_k+} gives $2(1+2c)\CR{\max(T_j-1,1)}/(1-c) \ge \CR{T_{k^+}}$.
Finally, we have
\be\label{eq:ineq11}
 R^2(\max(T_j-1,1))
 = \frac{a}{\max(T_j-1,1)}
 = \frac{T_j}{\max(T_j-1,1)} \frac{a}{T_j}
 \le \frac{la}{\max(l-1,0.5)T_j}
 = (\rho\CR{T_j})^2,
\ee
which gives the result.

\medskip{\bf Case 2.}
$j=\hk^*$ and $j\ne k^*$:

Now we have
\begin{align*}
 \hDe_j
& = \hmu_j-\hmu_{\hk^+}\\
&\le \hmu_j-\hmu_{k^*}\\
&\le \mu_j-\mu_{k^*} + c\CR{T_{k^*}} + c\CR{T_j}\\
&\stackrel{\mathrm{(f)}}{\le} \De_j+2c(1+2c)\rho\CR{T_j}/(1-c) + c\CR{T_j}
 = \De_j + (2\rho(1+2c)/(1-c)+1)c\CR{T_j}.
\end{align*}

{\bf (f)}
By definition $\De_{k^*} \le \De_j$,
 thus \eqref{eq:induction_i_k*} gives $2(1+2c)\CR{\max(T_j-1,1)}/(1-c) \ge \CR{T_{k^*}}$.
The claim follows using \eqref{eq:ineq11}.

\medskip{\bf Case 3.}
$j\ne\hk^*$ and $j=k^*$:
Now we have
\begin{align*}
 \hDe_j
& = \hmu_{\hk^*} -\hmu_j\\
&\le \mu_{\hk^*}-\mu_{k^*} + c\CR{T_j} + c\CR{T_{\hk^*}}\\
&\stackrel{\mathrm{(g)}}{\le} -\De_{\hk^*} + 2c/(1-c)\left(\De_{\hk^*}-\De_j\right) + c\CR{T_j} + 2c(1+2c)\rho\CR{T_j}/(1-c)\\
&\le - (1-2c/(1-c))\De_{\hk^*} -2c/(1-c)\De_j + c\CR{T_j} + 2c(1+2c)\rho\CR{T_j}/(1-c)\\
&\stackrel{\mathrm{(h)}}{\le} \De_j + c\CR{T_j} + 2c(1+2c)\rho\CR{T_j}/(1-c)
 = \De_j + (2\rho(1+2c)/(1-c)+1)c\CR{T_j}.
\end{align*}

{\bf (g)}
\eqref{eq:induction_i_hk*} can be rearranged as
\be\label{eq:ineq12}
 c\CR{T_{\hk^*}} \le 2c/(1-c)\left(\De_{\hk^*}-\De_j\right) + 2c(1+2c)\CR{\max(T_j-1,1)}/(1-c).
\ee
The claim follows using \eqref{eq:ineq12} and \eqref{eq:ineq11}.

{\bf (h)} This passage is true when $0\le 2c/(1-c)\le 1$, i.e., when $0\le c \le 1/3$.

\medskip{\bf Case 4.}
$j\ne \hk^*$ and $j\ne k^*$:
Now we have
\be\label{eq:ineq13}
 \hDe_j
 = \hmu_{\hk^*} - \hmu_j 
 \le \mu_{\hk^*} - \mu_j + c\CR{T_j} + c\CR{T_{\hk^*}}.
\ee

If $\hk^* = k^*$, we may write \eqref{eq:ineq13} as
\begin{align*}
 \hDe_j
& = -\hmu_j+\hmu_{\hk^*}\\
&\le -\mu_j + \mu_{\hk^*} + c\CR{T_j} + c\CR{T_{\hk^*}}\\
&\le \De_j +c\CR{T_j} +c\CR{T_{\hk^*}}\\
&\stackrel{\mathrm{(i)}}{\le} \De_j + c\CR{T_j} + 2c(1+2c)\rho\CR{T_j}/(1-c)
 = \De_j + (2\rho(1+2c)/(1-c)+1)c\CR{T_j}.
\end{align*}
{\bf (i)}
By definition $\De_{\hk^*} = \De_{k^*} \le \De_j$,
 thus \eqref{eq:induction_i_hk*} gives $2(1+2c)\CR{\max(T_j-1,1)}/(1-c) \ge \CR{T_{\hk^*}}$.
The claim follows using \eqref{eq:ineq11}.

Now if $\hk^* \ne k^*$, we may write \eqref{eq:ineq13} as
\begin{align*}
 \hDe_j
& = \hmu_{\hk^*}-\hmu_j\\
&\le \mu_{\hk^*}-\mu_j + c\CR{T_j} + c\CR{T_{\hk^*}}\\
&\le \De_j - \De_{\hk^*} + c\CR{T_j} + c\CR{T_{\hk^*}}\\
&\stackrel{\mathrm{(j)}}{\le} (1-2c/(1-c))\left(\De_j-\De_{\hk^*}\right) + c\CR{T_j}+2c(1+2c)\rho\CR{T_j}/(1-c)\\
&\stackrel{\mathrm{(k)}}{\le} \De_j + c\CR{T_j} + 2c(1+2c)\rho\CR{T_j}/(1-c)
 = \De_j + (2\rho(1+2c)/(1-c)+1)c\CR{T_j}.
\end{align*}
{\bf (j)}
Again, this claim follows using \eqref{eq:ineq12} and \eqref{eq:ineq11}.

{\bf (k)} This passage is true when $0\le 2c/(1-c)\le 1$, i.e., when $0\le c \le 1/3$.
\end{proof}

\vskip 0.2in
\bibliography{MTLTransfer2025dec2,FL-MTL,ngbib,allocation}

\end{document}